# Detection of Adverse Drug Events in Dutch clinical free text documents using Transformer Models – benchmark study

## Authors

Rachel M. Murphy*, PhD [a,b,c]

Nishant Mishra*, MSc [a,d]

Nicolette F. de Keizer, PhD [a,b,c]

Dave A. Dongelmans, MD PhD [c,e]

Kitty J. Jager, MD PhD [a,c,f]

Ameen Abu-Hanna, PhD [a,d,g]

Joanna E. Klopotowska**, PharmD PhD [a,b,c] (corresponding author, j.e.klopotowska@amsterdamumc.nl)

Iacer Calixto**, PhD [a,d,g]

*shared first authors*

**shared last authors*

[a] Amsterdam UMC location University of Amsterdam, Department of Medical Informatics, Meibergdreef 9, Amsterdam, the Netherlands

[b] Amsterdam Public Health, Digital Health, Amsterdam, the Netherlands

[c] Amsterdam Public Health, Quality of Care, Amsterdam, the Netherlands

[d] Amsterdam Public Health, Methodology, Amsterdam, the Netherlands

[e] Amsterdam UMC location University of Amsterdam, Department of Intensive Care Medicine, Meibergdreef 9, Amsterdam, the Netherlands

[f] Amsterdam Public Health, Aging & Later Life, Amsterdam, the Netherlands

[g] Amsterdam Public Health, Mental Health, Amsterdam, the Netherlands

Word count

(excluding title page, abstract, references, figures, and tables)

4.766

# Abstract

## Objective

To set a benchmark for adverse drug event (ADE) detection in Dutch clinical free text documents using several transformer models, clinical scenarios and fit-for-purpose performance measures.

## Materials and methods

In this study anonymized clinical free text documents in form of clinical progress notes of patients admitted to intensive care unit (ICU) of one academic hospital and discharge letters of patients admitted to Internal Medicine wards to two non-academic hospitals were reused. We trained a Bidirectional Long Short-Term Memory (Bi-LSTM) model and four transformer-based Dutch and/or multilingual encoder models (BERTje, RobBERT, MedRoBERTa.nl, and NuNER) for the tasks of named entity recognition (NER) and relation classification (RC) using 102 richly annotated Dutch ICU clinical progress notes. We evaluated our ADE RC models internally using gold standard (two-step task) and predicted entities (end-to-end task). In addition, all models were externally validated on detecting ADEs at the document level. We report both micro- and macro-averaged F1 scores, given the imbalance of ADEs in the datasets.

## Results

Although differences for the ADE RC task between the models were small, MedRoBERTa.nl was the best performing model with macro-averaged F1 score of 0.63 using gold standard and 0.62 using predicted entities. The MedRoBERTa.nl models also performed the best in our external validation and achieved recall of between 0.67 to 0.74 using predicted entities, meaning between 67 to 74% of discharge letters with ADEs were detected.

## Discussion

We set a benchmark for ADE detection by reusing Dutch clinical free text documents, with MedRoBERTa.nl performing best overall, based on internal and external validation.

## Conclusion

Our benchmark study presents a robust and clinically meaningful approach for evaluating language models for ADE detection in clinical free text documents. Our study highlights the need to use appropriate performance measures fit for the task of ADE detection in clinical free text documents and envisioned future clinical use.

# Introduction

Adverse drug events (ADEs) are a major public concern [1] and one of the leading causes of preventable patient harm in hospitals [2-4]. Information about frequency, type, and patients concerned in ADEs is essential to monitor and continuously optimise medication safety in hospitalised patients. However, this information is predominantly documented by physicians in free text documents [5]. Retrieving ADE information would therefore require the manual inspection of a vast number of clinical free text documents, which is prohibitively time-consuming and not feasible for routine and organization wide ADE monitoring. Automating ADE detection using natural language processing (NLP) seems a promising direction for identifying ADEs to support healthcare professionals charged with improving medication safety [4, 6].

Based on both our scoping literature review on NLP models for ADE detection from clinical free text documents and more recent studies published since that review [6-8], we identified four key limitations of the existing NLP models for ADE detection and their evaluation: 1) performance measures used are often not adequate for imbalanced datasets like datasets containing ADEs where ADEs are rare in comparison to other entities (e.g., drugs or disorders), 2) many datasets represent ADEs as entities rather than relations between a drug and disorder leading to avoidable errors in classification, 3) none of the models have been externally validated, and 4) finally none of the NLP models have been trained and validated using Dutch clinical free text documents. In this study, we address these limitations.

Our main goal was to benchmark four different transformer-based encoder language models for the clinical use case of ADE detection in Dutch clinical free text documents, and in particular clinical notes and discharge letters. The majority of studies in ADE domain used Bidirectional Long Short-Term Memory (Bi-LSTM) models [6], and recently other language models have been investigated [8]. For that reason, we used a Bi-LSTM baseline in our experiments. We focus on transformer models as they have recently shown to outperform Bi-LSTM models in ADE detection in English clinical free text documents. We deliberately did not include generative (or decoder-only) language models to focus on task-specific models that can be effectively fine-tuned and adopted for real-world clinical use, ensuring reliability and applicability in Dutch healthcare settings. Furthermore, a study comparing two generative pre-trained transformers (GPTs) to BERT-based models for clinical named entity recognition (NER) related to ADEs in clinical notes showed that the BERT-based models outperformed the large GPTs while being computationally more efficient [9]. Furthermore, investigating ADE detection from clinical free text documents in non-English settings is crucial, given the significant bias toward English-language data [10]. With this benchmark, we hope to enable researchers with a better understanding of the strengths and limitations of commonly used methods when applied to different healthcare systems and to languages other than English.

Finally, we selected the Intensive Care Unit (ICU) setting for our benchmark study, as ICU patients experience a higher incidence of ADEs compared to those in other hospital wards [11]. Consequently, the need for optimised ADE detection from clinical free text documents is particularly urgent in the ICU, where ADEs can lead to severe complications in an already vulnerable patient population.

Our paper's four main contributions are:

1. To the best of our knowledge, we are the first to benchmark several transformer models to detect ADEs in Dutch clinical free text documents like clinical notes and discharge letters.

2. We externally validated the trained models using a dataset that differ from the primary dataset—i.e. the dataset on which the models were trained and tested—in document type, patient characteristics, and medical specialty.

3. We evaluated the models from two medication safety monitoring scenarios: 1) identifying clinical progress notes with ADEs for further review by clinicians, and 2) for quantifying drugs and disorders involved in ADEs, and quantifying ADEs themselves.

4. We evaluated the models using often overlooked performance metrics adequate for imbalanced datasets, while still including common metrics widely reported in the literature.

## Related work

Lack of NLP-based ADE detection models for Dutch clinical free text documents may be explained by the lack of fit-for-purpose corpora to train such models, until very recently [12]. Wasylewicz et al. used a rule-based text-mining approach to detect

ADEs in Dutch clinical notes, achieving a sensitivity of 57% with a positive predictive value (PPV) of 32% [13]. They further extended their rules with additional deduplication and terminology matching steps, improving the sensitivity to 73% and the PPV to 70%, with an F1 score of 0.71 [14].

In 2018, two shared task challenges included tasks on detecting ADEs in clinical notes written in English. Track 2 of the 2018 n2c2 challenge evaluated the tasks of entity recognition, relation extraction and end-to-end tasks for ADEs and medication attribute detection (dose, strength, form, and so forth) [15]. The MADE 1.0 challenge, which also ran in 2018, included an entity recognition task (with ADE as an entity label, along with medication name and medication attributes), a relation extraction task (medication-ADE relationship) and an end-to-end task [16]. For both challenges, Bi-LSTM models were the most frequently used (these challenges pre-dated the release of transformer models) [6].

Since these challenges, others have made use of their datasets to test novel methods [7, 17-19]. In particular, Mahendran et al. achieved strong results on the relation classification task with three transformer encoder models: BERT, BioBERT, and ClinicalBERT, which all achieved micro-averaged F1 scores of 0.97 on the ADE-Drug relation extraction task [8]. Yang et al. [20] also achieved state-of-the-art performance on the relation classification tasks for n2c2 2018 and MADE 2018 data using transformer models like BERT, RoBERTa and XLNet and their clinically pre-trained variants. They showed that framing the problem as binary classification works better than multi-class classification. They made use of cross sentence representations generated by these transformers. To the best of our knowledge, these aforementioned models were not externally validated.

# Methods

## Study setting

This study was conducted between 01-2024 and 07-2025 at Amsterdam University Medical Centre (Amsterdam UMC), a large academic medical centre in the Netherlands.

## Datasets

We used three datasets in our experiments. Two of the datasets come from ICU admissions with acute kidney injury (AKI) at Amsterdam UMC.

The primary dataset: the *Dutch ADE corpus* contains 102 de-identified clinical progress notes, each belonging to one ICU admission. The notes in this corpus were written by ICU physicians on the last day of the ICU admission and therefore contain clinical summary spanning the entire ICU stay of the patient. Each note contains mentions by ICU physician of at least one specific ADE: a drug-related AKI [12]. The Dutch ADE corpus includes a crucial annotation that allows us to make the evaluation of the transformer models more relevant to clinical practice. It includes a mapping from each single ADE relation annotated in the clinical progress note (consisting of mentions of one drug and one disorder entity) to one unique ADE group, where applicable. This is important since different ADE relations annotated in a clinical progress note of one patient may refer to the same ADE from a clinical stance, e.g., if the same single ADE is mentioned multiple times in different parts of the same note. Moreover, a unique ADE group may consist of multiple drugs suspected to cause one specific disorder, e.g. a physician suspects furosemide, gentamicin and aciclovir as drugs that may have caused AKI. In such a case, there are three ADE relation annotations (furosemide-AKI, gentamicin-AKI, aciclovir-AKI) mapped to one unique ADE group. In this study, the annotations in the ADE corpus are used as ground-truth annotations.

The second dataset: the *ICU AKI corpus*, contains de-identified notes of ICU patients both with and without ADEs, which we used to gain a more realistic idea of the performance of the models on unseen data which may or may not contain ADEs and having sparse annotations. None of the ADEs in this dataset are drug-related AKI (DAKI).

The third dataset: the *WINGS corpus,* was used for external validation and contains 100 de-identified discharge letters of internal medicine patients admitted to two non-academic Dutch hospitals and collected for the purpose of a multicentre study on clinical pharmacists' interventions to reduce ADEs [21]. Of these 100 de-identified discharge letters, 36 contained *at least one* ADE annotated and confirmed by a medical expert panel [21]. None of these ADEs are DAKI and also here the annotation was sparse on document level. For the main characteristics of the datasets, see figure 1.

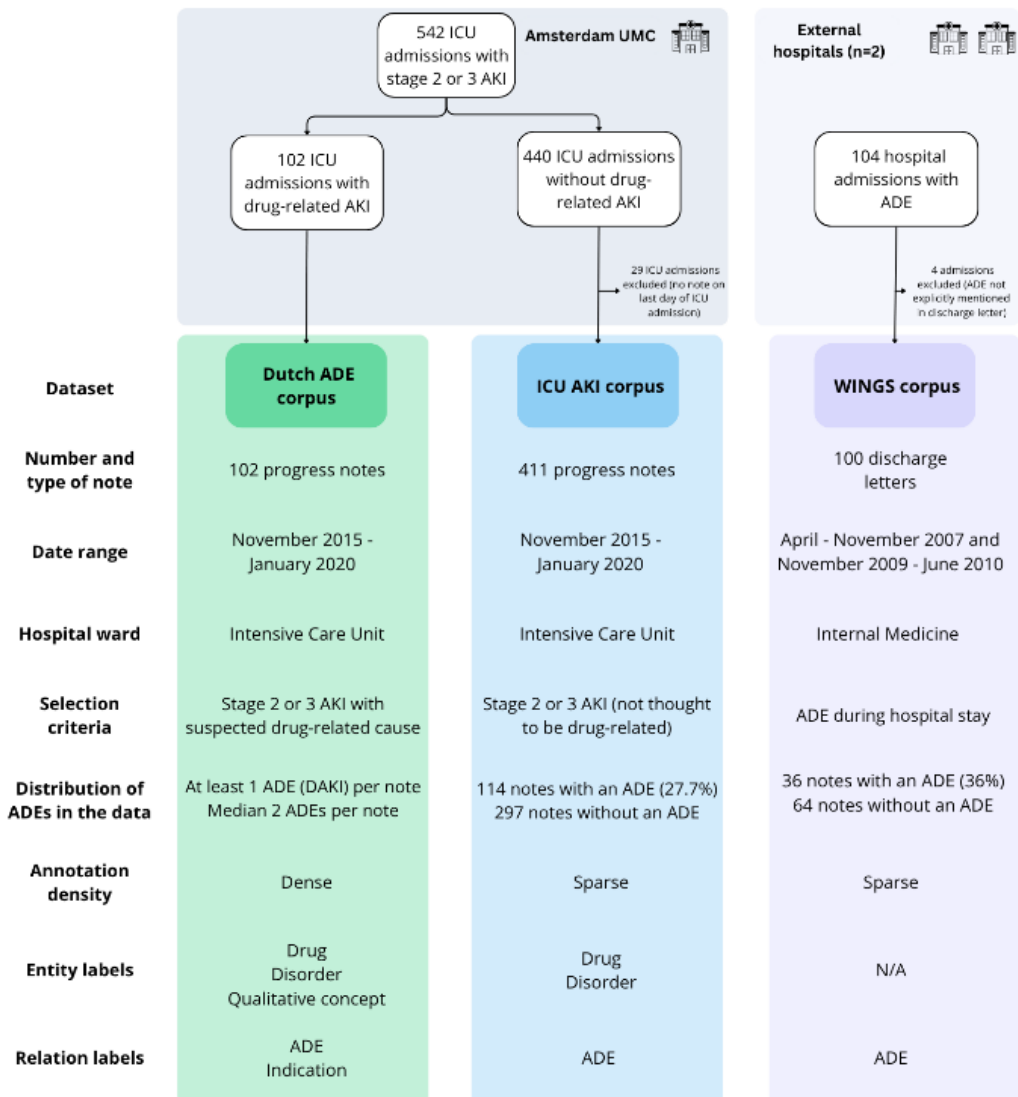

**Fig 1** Description of the three datasets used in the experiments. ADE: adverse drug event. AKI: acute kidney injury. DAKI: drug-related acute kidney injury. ICU: intensive care unit. Dense annotation: all entities and relations in the note are annotated. Sparse annotation: only entities relevant to the ADE mention and the ADE relations are annotated.

## Tasks

### Named Entity Recognition

Named entity recognition (NER) is a task where the goal is to locate and classify specific named entities (concepts) in free text. In the context of this study, NER is about identifying the mentions of 'drugs' and 'disorders' appearing in the clinical progress notes.

### Relation Classification

The goal in the relation classification (RC) task is to identify relationships between pairs of entities in free text. It assumes the existence of at least two entities, either available as ground-truth annotations or predicted by a NER model. In the context of this study, given a pair of drug and disorder entities in a clinical progress note, RC is about predicting whether the pair is an 'ADE', a 'drug indication', or neither.

## Models

Please see Supplement 1 for more details about the models, task, and hyperparameter tuning.

### Bi-LSTM model as baseline

We trained a Bidirectional Long Short-Term Memory (Bi-LSTM) network, which is a type of Recurrent Neural Network architecture that has shown robust performance across ADE-related NLP tasks [6]. We embedded input tokens using fastText Dutch word embeddings [22] and fed these the LSTM, which then predicted per-token NER labels. We used 300-dimensional FastText word embeddings and a Bi-LSTM with a hidden size of 256.

### Transformer models for benchmarking

Transformer-based models produce contextualised token embeddings and are among the state-of-the-art for NER and RC tasks [8, 23, 24]. One of the most popular models in use is the Bidirectional Encoder Representations from Transformers (BERT). We used different fine-tuned Dutch language versions of BERT to label entities for NER tasks: BERTje [25], RobBERT [26], MedRoBERTa.nl [27], and NuNER-multilingual [28], (see figure 2).

| | BERTje | RobBERT | MedRoBERTa.nl | NuNER-multilingual |
|---|---|---|---|---|
| Base model | BERT | RoBERTa | RoBERTa | BERT |
| Language | Dutch | Dutch | Dutch | Multilingual |
| Domain | General | General | Medical | General |
| Pre-training data | Books, TwNC (news), SoNaR-500, Web news, Wikipedia | Dutch section of OSCAR corpus (web crawl) | Electronic health records | OSCAR corpus (web crawl) |
| Number of parameters | 108 million | 110 million | 110 million | 125 million |

**Fig 2** Comparison of main model features. TwNC: Twente News Corpus, a corpus of Dutch language newspapers, television subtitles, news transcripts, among other things. SoNaR-500: corpus of more than 500 million words of text from various domains. OSCAR: Open Super-large Crawled Aggregated coRpus, a multilingual open-source corpus containing data from the web.

## Conditional Random Fields

Conditional random fields (CRF) are statistical models that model dependencies between output labels, which is important for token classification tasks where the label space has structure, such as NER [29, 30]. For that reason, we combined the Bi-LSTM and Transformer models with a linear chain CRF for the NER task.

## Experimental set up

For the Dutch ADE corpus, we used 5-fold cross validation with 60/20/20 train/validation/test percentual splits and performed hyperparameter tuning on the first fold for NER (since this already gave us stable results) and across all folds for RC. Figure 3 represents a simplified overview of our method, while Figure 4 provides a more detailed visual explanation of the experimental set up. Full details of the method are presented in Supplement 1.

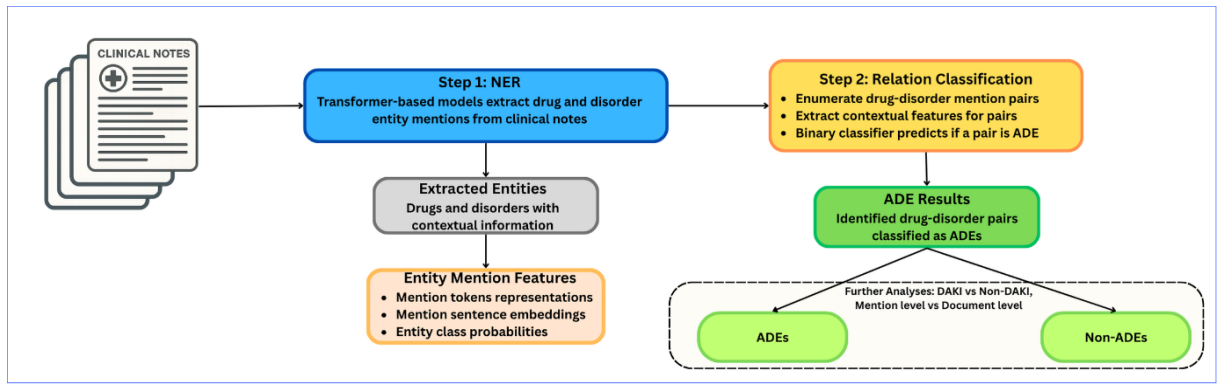

**Fig 3** Simplified overview of the method.

Our experimental setup consisted of three parts:

*Part 1 (the data):* We have an input consisting of sentences with gold-standard entity labels and two gold-standard binary relation labels (prescribing indication and adverse drug event or ADE). Entity labels follow the begin-inside-outside (BIO) format [ADD CITATION] with tags: B(drug), I(drug), B(disorder), I(disorder), O [31]. In short, B tags are used to denote the beginning of a named entity, and I tags are used inside a named entity with multiple words.

*Part 2 (the Entity Classification or EC model):* Given the input from part 1, the EC model predicts probabilities for (BIO) output labels for each token.

*Part 3 (the Relation Classification or RC model):* The RC model consists of a simple 4-layer multi-layer perceptron. The prediction of each relation is framed as a binary classification problem. We collect contextualised word embeddings for the CLS token (i.e., *context embedding*) and for the two candidate entities from the last layer of the EC Transformer model (i.e., *entity embeddings*). Based on the entity labels predicted by the EC model, we generate all *drug-disorder* pair combinations where drug-disorder mentions are at most 4 sentences apart in the input. For each pair, we create a feature vector with: the context embedding (2*768 dimensions) obtained from CLS token representation in the EC model; the entity embeddings (2*768 dimensions) obtained from drug and disorder mentions in the EC model; and the label probability vectors (2*8 dimensions) for drug and disorder entities as predicted by the EC model. This 3,088-dimensional feature vector that represents the entity pair is then passed to two binary relation classification models (each a simple 4-layer Multi-Layer Perceptron): one model predicts ADE (vs. non-ADE) and the other predicts prescribing indication (vs. no prescribing indication). The dimensions of the feature vector used for RC with baseline Bi-LSTM models were different and included: FastText embedding vectors (4*300 dimension) and LSTM hidden vectors (4*256 dimension)—in both cases, two for each of the entities and two for their context representation—, along with (2*5 dimension) output probability vectors for the entities predicted. Individual features are concatenated into a 2,034-dimensional feature vector for relation classification. Please refer to Supplement 1 for detailed information about how we performed hyperparameter search and tuning for all tasks.

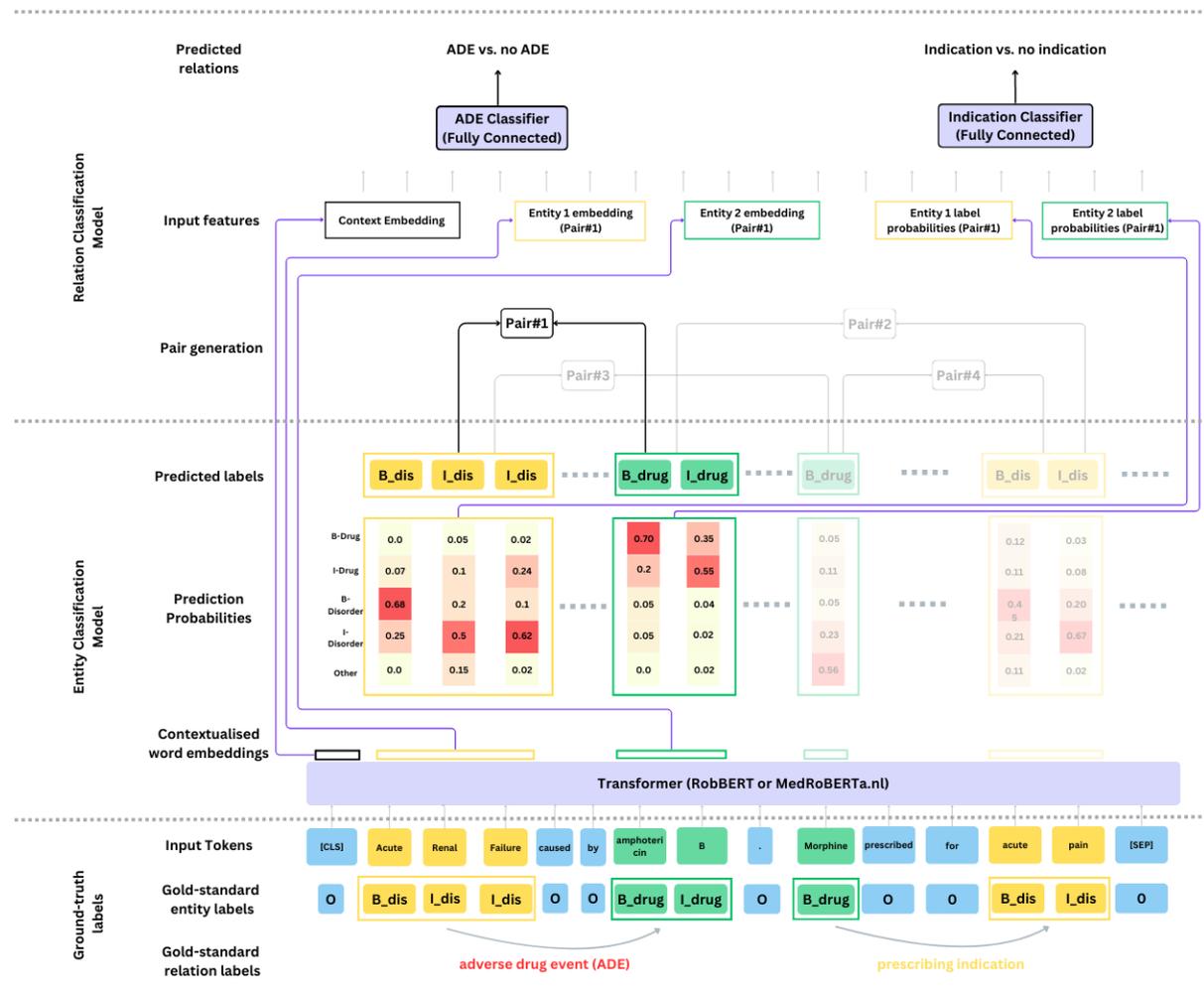

**Fig 4** Diagram illustrating the main components of our method consisting of three parts (should be read from the bottom to the top).

# Tasks, evaluation, and performance measures

## Tasks

In this work, we tackle four tasks:

1. Predicting drug and disorder entities (NER),
2. Predicting densely annotated ADE and prescribing indication relations (RC) using gold standard entities as input (two-step task),
3. Predicting densely annotated ADE and prescribing indication relations (RC) using predicted entities as input (end-to-end task),
4. Predicting if a note has at least one mention of an ADE relation (RC) (end-to-end task at the document level).

## Dutch ADE corpus evaluation

We evaluated the Bi-LSTM and transformer models on tasks 1, 2, and 3 using the Dutch ADE corpus. For all tasks, we calculated micro- and macro-averaged F1 score using the python scikit-learn library [32]. Macro-averaging shows the performance across classes treating each class as equally important, whereas micro-averaging gives equal weight to every instance and shows average performance across all predictions. Micro-averaging can hide low performance on minority classes for classification problems with class imbalance. We calculated both strict and lenient entity matching. In strict matching, we require an exact span match between predicted and gold-standard entities. In lenient matching, we require an overlap by at least one token between predicted and gold-standard entities' spans. Moreover, we plotted the precision-recall (PR) curve for each ADE RC model on the internal validation set of the Dutch ADE corpus to provide more details on the performance trade-off of the models.

We evaluated our RC models at the level of *single ADE relation* and at the level of the *unique ADE group*. Evaluation at the level of a single ADE relation is the standard evaluation setting used in all previous works [8, 15, 16] where each drug-disorder pair with an ADE annotation are taken individually. To evaluate RC at the unique ADE group level, we defined success in two ways: 1) *at least one* ADE relation part of the unique ADE group was identified (henceforth referred to as the 'easy' setting), and 2) *all* ADE relations that are part of the same unique ADE group were identified ('hard' setting). Evaluation under the easy setting can be closer to one of the likely intended clinical uses of such a model for medication safety monitoring where the model is used to scan clinical free text documents to identify ADEs mentioned by physicians and pass these free text documents for further manual causality assessment by clinical pharmacists and structured registration of the confirmed ADEs in the EHR system. Evaluation under the hard setting is more interesting from a methodological point-of-view, since it pinpoints to what extent methods can reliably identify every single ADE relation that compose a unique ADE group. Furthermore, this evaluation also provides clues about the ability of the models to quantify drugs and events involved in ADEs and ADEs themselves also for the purpose of medication safety monitoring in hospitals.

### ICU AKI corpus and WINGS corpus evaluation

In both the ICU AKI and the WINGS dataset, we evaluated Bi-LSTM and transformer models on the task of the end-to-end ADE prediction at the document level (task 4). Here, we used entity mentions predicted by the NER model, i.e., we do not use ground-truth drug and disorder spans and labels. Moreover, each note in the ICU AKI corpus and in the WINGS corpus has a single (binary) ADE label denoting whether an ADE relation exists between any drug-disorder pair in the clinical note. Concretely, for each note we predict the presence or absence of an ADE relation in a document.

### Threshold selection

We used the precision-recall curve on the Dutch ADE corpus validation set to search for F1 and F2 scores at different thresholds and find the best to use for inference on unseen test data (across all datasets). The F2 score is an F-score where recall is weighted twice as importantly as precision compared to the F1 score. In the context of ADE detection from clinical free text documents using NLP, the F2 score is particularly useful when it is more important to prioritise recall over precision. This could be crucial in medical applications where missing ADEs mentioned in clinical free text documents can lead to serious consequences.

# Results

### Dutch ADE corpus

For NER, MedRoBERTa.nl achieved the highest micro-averaged F1 score in all settings, e.g., overall vs. drugs vs. disorders, strict vs. lenient matching (see Table

S2.1 in Supplement 2). On average, the drug class achieved higher micro-averaged F1 scores than the disorder class (average across the four models of 0.8927 drug vs. 0.7336 disorder micro-averaged F1 score). For micro-averaged precision, recall and F1 scores on validation set, see table S2.2 in Supplement 2.

For RC, there was little variation between the models in terms of micro-averaged F1 score on the ADE RC task using predicted entities (end-to-end task, range 0.9846 - 0.9917). The macro-averaged F1 scores for the ADE RC task were much lower (0.5812 - 0.6222). For the ADE RC task using gold standard entities, the micro- and macro-averaged F1 scores followed similar patterns (see Table 1). We provide additional experiments including validation set results in Supplement 2.

When predicting at the ADE group level, the results are more nuanced. Although MedRoBERTa.nl obtains the highest scores on average across folds, when accounting for the standard deviation of the scores there is no clear-cut best-performing model. Moreover, standard deviations in Table 2 are an order of magnitude higher than, for instance, those reported in Table 1. We note that performing model selection using the F2 score and using that same metric to choose the classification threshold leads to considerably higher ADE retrieval performance. Figure 5 depicts the PR curve for each ADE RC model on the internal validation set. MedRoBERTa.nl has the highest area under the curve while BERTje is consistently one standard deviation below other models.

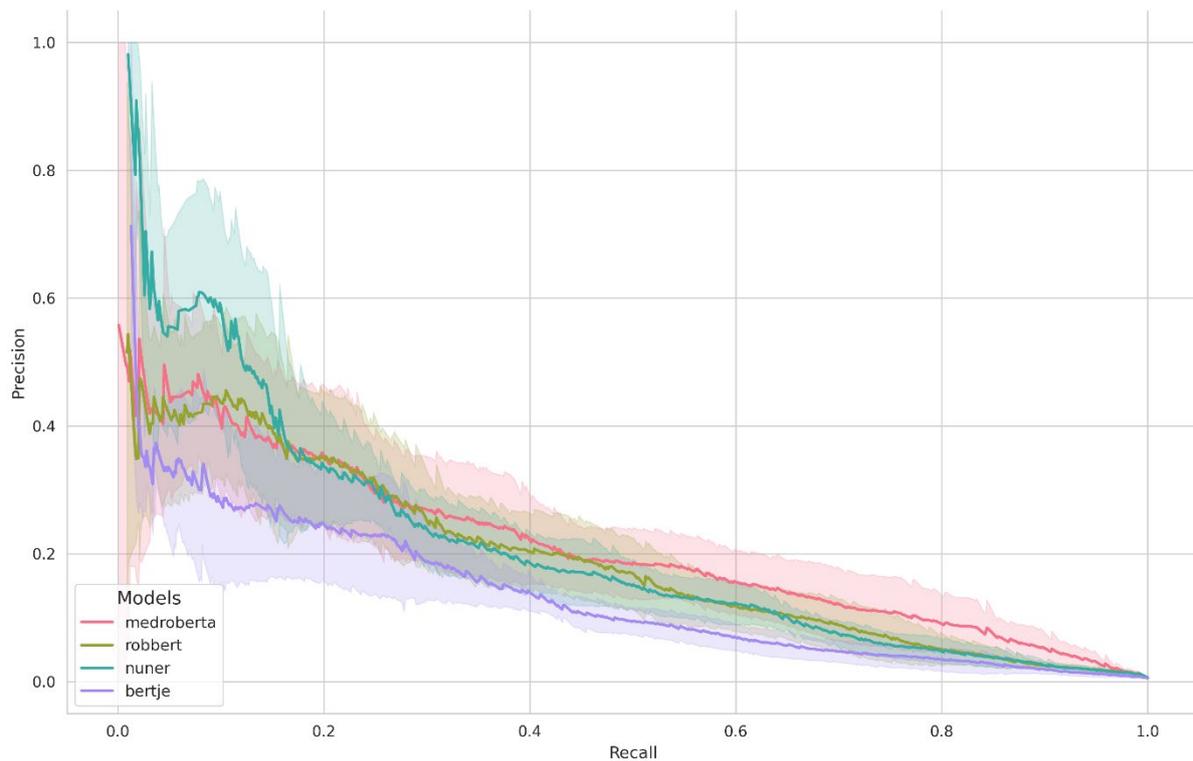

**Fig 5** Mean recall (sensitivity) versus mean precision (positive predictive value) for the adverse drug event class over five folds with one standard deviation on the Dutch ADE corpus.

## ICU AKI corpus

On the ICU AKI corpus, MedRoBERTa.nl demonstrated the best performance when selecting a threshold based on F1 score, while results were a bit more mixed when selecting a threshold based on F2 score (e.g. with MedRoBERTa.nl, RobBERT or BERTje performing best depending on the metrics considered).

**Table 1** Performance measures for relation classification models on adverse drug event and indication classes on the test set of the Dutch ADE corpus (results correspond to tasks 2 and 3).

| Model | Micro-averaged F1 score* | | Macro-averaged F1 score* | |
|---|---|---|---|---|
| | Adverse Drug Event | Indication | Adverse Drug Event | Indication |
| **Using gold standard entities—two-step task (task 2)** | | | | |
| Bi-LSTM | **0.9896 ± 0.0039** | 0.9513 ± 0.0138 | 0.5864 ± 0.0227 | 0.5373 ± 0.0042 |
| BERTje | 0.9880 ± 0.0039 | 0.9680 ± 0.0045 | 0.6005 ± 0.0146 | 0.5925 ± 0.0135 |
| RobBERT-base | 0.9879 ± 0.0030 | **0.9744 ± 0.0037** | 0.6181 ± 0.0271 | 0.6105 ± 0.0168 |
| MedRoBERTA.nl | 0.9847 ± 0.0049 | 0.9691 ± 0.0070 | **0.6301 ± 0.0196** | **0.6202 ± 0.0110** |
| NuNER-multilingual | 0.9892 ± 0.0034 | 0.9700 ± 0.0045 | 0.6133 ± 0.0315 | 0.6082 ± 0.0131 |
| **Using entities predicted with lenient matching—end-to-end task (task 3)** | | | | |
| Bi-LSTM | **0.9917 ± 0.0022** | 0.9474 ± 0.0095 | 0.5812 ± 0.0287 | 0.5517 ± 0.0179 |
| BERTje | 0.9846 ± 0.0072 | 0.9694 ± 0.0036 | 0.5963 ± 0.0160 | 0.5960 ± 0.0100 |
| RobBERT-base | 0.9884 ± 0.0024 | 0.9738 ± 0.0043 | 0.6040 ± 0.0193 | 0.6088 ± 0.0106 |
| MedRoBERTA.nl | 0.9891 ± 0.0025 | 0.9728 ± 0.0063 | **0.6222 ± 0.0094** | **0.6180 ± 0.0120** |
| NuNER-multilingual | 0.9896 ± 0.0026 | **0.9753 ± 0.0013** | 0.6041 ± 0.0251 | 0.6173 ± 0.0116 |

*  with one standard deviation, 5-fold cross-validation

**Table 2** Predictions of adverse drug event group on the test set of the Dutch ADE corpus (5-fold cross-validation).

| Model | Easy evaluation* | | Hard evaluation** | |
|---|---|---|---|---|
| | % of ADE groups predicted | % of ADE groups predicted | % of ADE groups predicted | % of ADE groups predicted |

|  | (best F1 threshold) | (best F2 threshold) | (best F1 threshold) | (best F2 threshold) |
|---|---|---|---|---|
| B-iLSTM | 21.68 ± 4.82 | 32.44 ± 6.01 | 12.86 ± 5.51 | 22.41 ± 5.32 |
| BERTje | 35.67 ± 9.11 | 49.17 ± 3.74 | 28.26 ± 10.16 | 37.15 ± 6.12 |
| RobBERT-base | 35.65 ± 7.41 | 47.62 ± 6.55 | 25.41 ± 10.37 | 35.07 ± 7.50 |
| MedRoBERTA.nl | **37.24 ± 6.74** | **60.41 ± 4.68** | **28.47 ± 9.20** | **49.53 ± 10.46** |
| NuNER-multilingual | 30.90 ± 8.28 | 51.58 ± 5.18 | 21.30 ± 9.07 | 41.42 ± 6.53 |

ADE: adverse drug event. *Easy evaluation refers to detection of any single ADE relation belonging to ADE group. ** Hard evaluation refers to the detection of all ADE relations belonging to unique ADE group.

**Table 3** Identification of the presence or absence of an adverse drug event using the ICU AKI corpus at the document level (results correspond to Task 4).

|  | F1 threshold | | | |
|---|---|---|---|---|
| Model | Precision | Recall | F1 score | Specificity |
| BERTje | 0.617 ± 0.017 | 0.621 ± 0.028 | 0.567 ± 0.078 | 0.587 ± 0.206 |
| RobBERT-base | 0.627 ± 0.029 | 0.600 ± 0.017 | 0.600 ± 0.014 | 0.833 ± 0.074 |
| MedRoBERTA.nl | **0.649 ± 0.016** | **0.627 ± 0.021** | **0.627 ± 0.013** | **0.828 ± 0.079** |
| NuNER-multilingual | 0.615 ± 0.012 | 0.598 ± 0.025 | 0.597 ± 0.027 | 0.826 ± 0.068 |
|  | F2 threshold | | | |
| Model | Precision | Recall | F1 score | Specificity |
| BERTje | **0.602 ± 0.015** | 0.593 ± 0.012 | 0.467 ± 0.050 | 0.319 ± 0.097 |
| RobBERT-base | 0.579 ± 0.014 | 0.595 ± 0.016 | **0.555 ± 0.031** | **0.569 ± 0.095** |
| MedRoBERTA.nl | **0.602 ± 0.012** | 0.620 ± 0.020 | 0.539 ± 0.036 | 0.454 ± 0.068 |
| NuNER-multilingual | 0.589 ± 0.010 | 0.605 ± 0.013 | 0.535 ± 0.037 | 0.474 ± 0.093 |

**Table 4** Identification of the presence or absence of an adverse drug event using the WINGS corpus at the document level (results correspond to Task 4 - External validation).

|  | F1 threshold | | | |
|---|---|---|---|---|
| Model | Precision | Recall | F1 score | Specificity |
| BERTje | 0.697 ± 0.052 | 0.633 ± 0.035 | 0.622 ± 0.046 | 0.822 ± 0.183 |
| RobBERT-base | 0.698 ± 0.024 | 0.578 ± 0.025 | 0.552 ± 0.041 | 0.956 ± 0.012 |
| MedRoBERTA.nl | **0.794 ± 0.043** | **0.674 ± 0.032** | **0.681 ± 0.038** | **0.959 ± 0.025** |
| NuNER-multilingual | 0.683 ± 0.013 | 0.579 ± 0.023 | 0.555 ± 0.044 | 0.941 ± 0.023 |
|  | F2 threshold | | | |
| Model | Precision | Recall | F1 score | Specificity |
| BERTje | 0.661 ± 0.024 | 0.659 ± 0.024 | 0.649 ± 0.038 | 0.706 ± 0.128 |
| RobBERT-base | 0.692 ± 0.024 | 0.637 ± 0.034 | 0.638 ± 0.040 | 0.891 ± 0.026 |
| MedRoBERTA.nl | **0.736 ± 0.021** | **0.736 ± 0.021** | **0.736 ± 0.021** | **0.806 ± 0.025** |
| NuNER-multilingual | 0.698 ± 0.024 | 0.680 ± 0.026 | 0.680 ± 0.026 | 0.809 ± 0.069 |

### WINGS corpus

On the WINGS corpus, MedRoBERTa.nl outperforms all other models in terms of predicting ADEs at the clinical note level by a large margin—i.e., 0.681 vs. 0.622 F1 achieved by the second-best model BERTje when using an F1 threshold, or 0.736 vs. 0.680 F1 achieved by the second-best model NuNER-multilingual using a F2 threshold (see table 4).

# Discussion

## Main findings

The aim of this study was to set a benchmark for the task of ADE detection in Dutch clinical free text documents using common models, including Bi-LSTMs and encoder-only general-domain and clinical transformer models. We designed our experiments with the goal to improve upon previous studies in terms of performance measures used, to carry on our evaluation with envisioned future clinical uses in mind, and by addressing the lack of external validation. For this purpose, we used three datasets. On the Dutch ADE corpus, all models investigated in this study achieved micro-averaged F1 scores of ± 0.98 for ADE RC task using both gold standard and predicted entities. Transformer-based models achieved macro-averaged F1 scores between 0.60 and 0.63 using gold standard and between 0.59 and 0.62 using predicted entities, whereas Bi-LSTMs slightly underperform compared to the least performant transformer models. Although the differences between the models were small, MedRoBERTa.nl was the best performing model for ADE RC in terms of macro-averaged F1 score. For the prediction at the level of single ADE relation within an ADE group (easy evaluation), MedRoBERTa.nl showed the best performance (% of ADE groups predicted with best F1 threshold of 37.2 ± 6.7, and best F2 threshold 60.4 ± 4.7). This means MedRoBERTa.nl was able to detect 37% of the ADEs in the Dutch ADE corpus, rising to 60% when recall was prioritised. This performance could be considered modest if contemplating implementation in the clinical setting, although it is still a great improvement on the current proportion of ADEs reported in a structured manner (estimated between 6 and 33%) [33, 34]. When we require all mentions in a unique ADE group to be predicted correctly to count as success (hard evaluation), MedRoBERTa.nl was the best performing model (% of ADE groups predicted with best F1 threshold of 28.5 ± 9.2, and best F2 threshold 49.5 ± 10.5). This means that MedRoBERTa.nl could retrieve the full ADE mention with all implicated drugs and any duplicate mentions in 28% of cases, rising to 49% when recall was prioritised. This shows the proportion of ADEs that could potentially be detected and recorded in an automated ADE monitoring system.

On the ICU-AKI corpus, models achieved moderate performance with F1 scores in the range 0.57 to 0.63 for the F1 threshold and in the range of 0.47 to 0.56 for the F2 threshold, with MedRoBERTa.nl outperforming all other models according to most metrics. Models also recorded consistently high specificity (0.82-0.83), with BERTje performing significantly worse than other models according to this metric (0.58). On the WINGS corpus, MedRoBERTa.nl again outperformed all the models with F1 score of 0.68 using F1 threshold and specificity of 0.96 for ADE RC using predicted entities at the document level. BERTje also performed well with a F1 score of 0.622 but had a comparatively lower specificity of 0.83, suggesting higher false positive rate.

In our ADE RC evaluation on the Dutch ADE corpus, we found that the choice of metric is very important. Looking at micro-averaged F1 score indicates very high performance for all models; however the micro-averaged F1 score strongly favours the majority class, whereas ADEs account for only 0.64% of the entity pairs used to train the RC model. Therefore, macro-averaged F1 scores and precision-recall curves give a more realistic assessment of the performance of the models in identifying rare outcomes such as ADEs. If we instead inspect the macro-averaged F1 scores, MedRoBERTa.nl is the best performing model by a reasonable margin. If we only look at the micro-averaged F1 scores, the Bi-LSTM is the best-performing model, although by a very small margin. The macro-averaged F1 results suggest that there is still room for improvement of these models.

The fact that MedRoBERTa.nl tends to outperform the other three models using ADE corpus could be explained by the fact that approximately half of the notes in the Dutch ADE corpus may have been included in the MedRoBERTa.nl *pre-training data*. The

MedRoBERTa.nl training data included clinical notes written at Amsterdam UMC in 2017 and 2018, and the Dutch ADE corpus includes notes written at Amsterdam UMC between November 2015 and January 2020. Therefore, data leakage may have occurred, giving MedRoBERTa.nl an unfair advantage. External validation of the models, however, shows that MedRoBERTa.nl obtains competitive results also on external corpora or, in other words, MedRoBERTa.nl did not overfit. Its high performance is most likely due to its training data (Dutch EHR clinical notes) being much more similar in structure and content than the general language used to train the other models.

Besides observing notable differences in model performance depending on the performance metric used, we also observed that the models' performance varies depending on the training objective as well as model selection. If we put more emphasis on recall—by using the F2 metric for model selection as well as threshold selection, the model gives better performance on the ADE group level results (table 2) as well as in the WINGS corpus (table 4), since these are recall-oriented experiments. On the flip side it can lead to poor PPV and make it less suitable for applications requiring lower false positive rates.

Using micro-averaged F1 scores can lead to uninformative results when comparing models for a task with such data imbalance as ADE detection, since the majority class performance dominates the micro-averaged F1 scores. Based on these findings, we suggest reporting macro-averaged F1 scores alongside micro-averaged F1 scores, as it allows for a better assessment of model performance for ADE identification in clinical free text documents, as well as enabling comparison between studies. We also recommend using F2 scores for threshold selection when dealing with ADE detection in an imbalanced dataset where it is important not to miss ADEs.

In our internal validation using the ICU AKI corpus and our external validation using the WINGS corpus, using the right metric for the model and threshold selection for downstream task is again crucial. Overall, MedRoBERTa.nl outperformed all other models by a moderate margin on the Dutch ICU corpus, and by a medium-to-large margin in the WINGS corpus. Interestingly, in the ICU AKI corpus, MedRoBERTa.nl trained using an F1 threshold obtained (marginally) better recall than when trained using an F2 threshold (0.627 vs. 0.620). We observed the opposite when applying

MedRoBERTa.nl on the WINGS corpus, in which case training using F2 threshold lead to considerably better recall than using F1 threshold (0.736 vs. 0.674). Overall, our results suggest that choosing the right model and the right metric—to use for model and threshold selection—can be non-trivial and can also affect the downstream generalisation of the model.

## Strengths and limitations

The strengths of this paper include the fact that this is, to the best of our knowledge, the first study in which Bi-LSTMs and various transformer models have been benchmarked for the task of detecting ADEs in Dutch ICU progress notes. Another important strength of our study is that we conducted external validation of our models. To the best of our knowledge, the results from external validation of an NLP model for identification of ADEs in EHR clinical free text documents have not yet been published. Another strength is the efficiency of our RC method in comparison to related methods that use transformers for RC [20], as we did not train or fine-tune a separate transformer model for RC, but instead reused the embeddings generated by the NER models for our entity pairs. Furthermore, we used both micro- and macro-averaged F1 scores for more realistic performance evaluation, and in addition F1 and F2 scores for threshold selection, as recall is considered more important so as not to miss ADE mentions. Lastly, we evaluated our models from two clinical perspectives in order to account for possible future uses for medication safety monitoring in clinical practice.

The limitations of this work include the fact that the gold standard training corpus is small, containing 102 clinical notes and just over 600 ADE-related drug-disorder pairs. Lastly, every note in the ADE corpus contained at least one ADE, which was not the case in the ICU AKI and WINGS corpus. Therefore, ICU AKI and WINGS datasets better reflect the pattern of ADE mentions in real world clinical free text documents, which will also contain negative samples (notes with no ADE mentions). We aim to design models better tailored to this scenario, where clinical notes and discharge letters are likely to have no ADEs.

## Conclusions and future research

Our benchmark provides a baseline for future research on ADE detection in Dutch clinical free text documents, and in particular clinical notes and discharge letters. We externally validated a Bi-LSTM and four transformer-based NLP models for ADE identification and aim to follow this up with clinical validation to assess whether these models are suitable for deployment in the clinical environment. We strongly advise to use macro-averaged F1 and F2 scores, as well as precision-recall curves in similar evaluations as these give a more realistic assessment of the performance of the models in detecting ADEs. In future work we also intend to train and benchmark larger general purpose and specialised models (with billions of parameters) for ADE detection to investigate their capabilities for ADE detection from clinical free text documents like clinical notes and discharge letters.

## Funding

This study was funded partly by Innovation fund 2019 of Amsterdam UMC (project number: 23088) and by The Netherlands Organization for Health Research and Development (ZonMw project number: 848018004). The funders had no role in the design of the study, the collection, analysis, and interpretation of data or in writing the manuscript. IC and NM are part of the project CaRe-NLP with project number NGF.1607.22.014 of the research programme NGF - AiNed Fellowship which is (partly) financed by the Dutch Research Council (NWO).

## Ethical considerations

This study was exempted from requiring ethics approval on 03/06/2019 (non-WMO waiver W19_207 # 19.252) by the Medical Ethics Committee of the Amsterdam University Medical Centre, location University of Amsterdam, The Netherlands, as it did not fall within the scope of the Dutch Medical Research Involving Human Subjects Act (WMO). Prior to obtaining the data from Research Data Management office of Amsterdam UMC, we performed a Data Protection Impact Assessment which was assessed and approved by the privacy officer of Amsterdam UMC. In accordance with Dutch and Amsterdam UMC regulation regarding reuse of routine care data for research, informed consent of patients is not required for anonymised data. Patients who do not agree to the reuse of their routine care data for research are excluded from data extractions by the Research Data Management office.

## Competing interests

No competing interest is declared.

# Data availability

The Dutch ADE corpus was generated as part of a previous study and will be shared on reasonable request to the corresponding author. The ICU-AKI corpus and the WINGS corpus cannot be shared to preserve the privacy of the individuals involved.

# Author contributions statement

Rachel M. Murphy: Conceptualisation, Data Curation, Methodology, Project Administration, Visualisation, Writing – Original Draft Preparation, Writing – Review & Editing

Nishant Mishra: Formal Analysis, Methodology, Software, Visualisation, Writing – Original Draft Preparation, Writing – Review & Editing

Nicolette F. de Keizer: Funding Acquisition, Supervision, Writing – Review & Editing

Dave A. Dongelmans: Funding Acquisition, Supervision, Writing – Review & Editing

Kitty J. Jager: Funding Acquisition, Writing – Review & Editing

Ameen Abu-Hanna: Funding Acquisition, Methodology, Supervision, Writing – Review & Editing

Joanna E. Klopotowska: Conceptualisation, Funding Acquisition, Methodology, Project Administration, Resources, Supervision, Writing – Review & Editing

Iacer Calixto: Conceptualisation, Formal Analysis, Methodology, Resources, Software, Supervision, Writing – Original Draft Preparation, Writing – Review & Editing

# Acknowledgments

We thank Rosa J. Jongeneel and Christiaan H. Koster for their annotation work of the Dutch ADE corpus, Rudy Scholte and Oscar van der Meer from Research Data Management of Amsterdam UMC for their assistance in data retrieval and advice on computing facilities for high-sensitive data, and Michel Paardekooper, privacy officer at Amsterdam UMC, for his advice and support related to working with high-sensitive data and de-identification procedure.

This study was part of the project Research in clinical prediction models and natural language processing with deep learning (project number NWO-2024.015) of the research programme Computing Time on National Computer Facilities. The computational resources used were financed by the Dutch Research Council (NWO).

events in hospitalized adults. J Gen Intern Med 1993;8(6):289-94 doi:10.1007/bf02600138.

# Detection of Adverse Drug Events in Dutch clinical free text documents using Transformer Models – benchmark study

# Supplement 1: Methods

## Authors

Rachel M. Murphy*, PhD [a,b,c]

Nishant Mishra*, MSc [a,d]

Nicolette F. de Keizer, PhD [a,b,c]

Dave A. Dongelmans, MD PhD [c,e]

Kitty J. Jager, MD PhD [a,c,f]

Ameen Abu-Hanna, PhD [a,d,g]

Joanna E. Klopotowska**, PharmD PhD (corresponding author, j.e.klopotowska@amsterdamumc.nl) [a,b,c]

Iacer Calixto**, PhD [a,d,g]

*shared first authors*

**shared last authors*

[a] Amsterdam UMC location University of Amsterdam, Department of Medical Informatics, Meibergdreef 9, Amsterdam, the Netherlands

[b] Amsterdam Public Health, Digital Health, Amsterdam, the Netherlands

[c] Amsterdam Public Health, Quality of Care, Amsterdam, the Netherlands

[d] Amsterdam Public Health, Methodology, Amsterdam, the Netherlands

[e] Amsterdam UMC location University of Amsterdam, Department of Intensive Care Medicine, Meibergdreef 9, Amsterdam, the Netherlands

[f] Amsterdam Public Health, Aging & Later Life, Amsterdam, the Netherlands

[g] Amsterdam Public Health, Mental Health, Amsterdam, the Netherlands

## Models

As we have Dutch language clinical free text documents, we used the following Dutch or multilingual models:

- **Bi-LSTM** is our baseline model which uses the Long Short Term Memory recurrrent architecture for sequential task. We use static word embeddings as inputs to the LSTM model as word/token representations, for which we used the fastText Dutch word embeddings, a 300-D embedding trained on the Commoncrawl corpus and Wikipedia.
- **BERTje** is a Dutch state-of-the-art transformer model. It is a BERT-based general purpose monolingual model trained exclusively on general Dutch language text data (including books, news articles, Wikipedia articles) [1]. It has 108M parameters.
- **RobBERT** is a large pre-trained general Dutch language model that can be fine-tuned to perform any text classification, regression, or token-tagging task [2]. It has 110M parameters. RobBERT is the current state-of-the-art Dutch RoBERTa-based transformer model.
- **MedRoBERTa.nl** is a RoBERTa-based transformer model pre-trained from scratch on nearly 10 million Dutch hospital notes sourced from Electronic Health Records of the Amsterdam UMC [3]. Since it is a domain-specific model trained on clinical data, it is meant to be used on medical NLP tasks in Dutch language settings. It also has 110M parameters.
- **NuNER-multilingual** is a compact transformer model specialised in NER that can be categorised as a task-specific foundation model [4]. It is created by pre-training a BERT-base model on a large NER dataset using a contrastive learning strategy with task specific encodings added during pre-training. It has a multilingual variant which was based on a multilingual. BERT-base we used in our experiments. It has 125M parameters.

## Experimental Design

### Named Entity Recognition

We trained NER models as a multi-class classifiers. We framed named entity (drug and disorder) recognition as a token classification task and used the BIO tagging scheme for labelling [5]. The BIO tagging scheme is a way of dividing entity labels according to the tokens used by the model (see figure S1.1 for a graphical representation).

**Sentence** Acute kidney injury caused by vancomycin and sepsis
**Tags** B-Disorder I-Disorder I-Disorder O O B-Drug O B-Disorder

**Fig. S1.1** Illustration of BIO tagging.

Relation Classification

We experimented with two options for the RC task: two-step and end-to-end RC. In two-step RC, we used pairs of gold-standard entities in the primary dataset as the inputs to the RC classifier (i.e. the entities as annotated by the medical experts). In the end-to-end RC, we used pairs of the predicted entities from our NER model as inputs to the RC classifier. The end-to-end RC is a more realistic setting, since at inference time reasonably only raw clinical notes are available. However, the two-step RC setting is still relevant since it tells us what we could obtain if we had a perfect NER model available.

We used a one-vs-all strategy and trained two binary classifiers: ADE vs. no-ADE and prescribing indication vs. no-prescribing indication. In order to generate candidate drug-disorder entity pairs: 1) we collected all drug entities and all disorder entities from a patient note, gold-standard or predicted; 2) we kept the drug-disorder pairs if these entities were within a maximum of 4 sentences apart in the note; 3) we extracted their embeddings and probabilities as discussed in detail in the *Relation Classification (RC) models* section below to train RC; 4) we applied SMOTE [6] to oversample positive pairs (ADE or indication); and 4) we randomly undersampled negative drug-disorder pairs to use as negative pairs.

We used 5-fold cross validation with 80/10/10 train/validation/test percentual splits and performed hyperparameter tuning on the first fold for NER, since using a single fold already gave us stable results, and across all folds for RC. We trained for a maximum of 10 epochs for NER and 100 epochs for RC, performed model selection according to its F1 score on the validation set, and used patience to stop training once a model did not improve for an optimal number of epochs (early stopping patience hyperparameter). The loss functions used for training are the per-token categorical cross-entropy (NER) and the binary cross-entropy-based focal loss (RC).

## Implementation Details

### Named Entity Recognition (NER) models

We fine-tuned Bi-LSTM, BERTje, NuNER-multilingual, MedRoBERTa.nl and RobBERT models. BERTje, NuNER-multilingual are BERT-based and MedRoBERTa.nl and RobBERT are RoBERTa-based. In all NER models we used Conditional Random Field (CRF) layers for prediction.

We trained all models using 5-fold cross-validation and conducted hyperparameter tuning for all models using the first fold. We trained for a maximum of 10 epochs and stopped training in case a model did not improve performance on the validation set after 3 epochs.

The hyperparameter search space was as follows:

- Learning rate: {1e-5, 2e-5, 3e-5, 4e-5, 5e-5, 1e-6}
- Max seq length: {128, 256, 512}
- Batch size: {8, 16, 32, 64}
- Training epochs: {8,10}
- Warmup ratio: {0.1, 0.2, 0.3}
- Early stopping patience: {2, 3, 4}

We used a learning rate warmup with reduceLROnPlateau to optimize model convergence. The best hyperparameter combination was: learning rate 3e-5, max seq len 512, batch size 8, training epochs 10, warmup ratio 0.2, and patience 3. In our model selection, we select the best-performing model according to the micro-F-1 score (with strict matching) on the validation set. We also experimented with different loss functions (i.e., focal loss, cross-entropy) and with and without CRF layer. We obtained the best results with cross-entropy loss using a CRF layer, and we report results only for those settings.

We trained the BiLSTM model in a 5-fold cross-validation setting, along with a CRF layer. Here we used word-level tokenization instead of the subword tokenizer that comes with transformers-based model, and we also used a pre-initialized static word embedding model for representation of input vectors. Similar to the transformer models, we performed extensive hyperparameter search across 50 trials, with the search space was:

- Number of Layers: {1,2,3}
- Hidden Dimension: {64, 128, 256, 512}
- Learning rate: [1e-2...1e-5]
- Batch size: {16, 32, 64}
- Weight Decay: [1e-6, 1e-1]
- Scheduler: {'reduce_on_plateau', 'cosine_annealing', 'one_cycle', 'exponential'}
- Early stopping patience: {3, 4,5 ,6}

## Relation Classification (RC) models

For RC, we used a simple multi-layer perceptron for classification. In both two-step and end-to-end RC the inputs to the RC classifier are: 1) the embeddings of pairs of drug and disorder entities (extracted with the NER model), 2) the embedding of the context in which these entities appear (i.e., the [CLS] token representation that gives an embedding for the entire sentence containing the entity, extracted with the NER model), and 3) the probability distribution over the labels assigned to the (drug and disorder) tokens by the NER models.

To train our RC models, we generated candidate entity (drug-relation) pairs and extracted embeddings for these candidate pairs. To generate candidate drug-relation pairs, we first extracted all combinations of drug-disorder entities from the ICU progress notes regardless of where these entities appear in the note. We only selected entity pairs that appeared within 4 sentences from each other. We chose 4 sentences as it covered 99.42% of all ADE relations in the gold-standard ADE corpus. In order to generate negative pairs, we only select drug-disorder pairs and not drug-drug or disorder-disorder since these pairs will never have a relation in our dataset context. We sampled a fixed ratio of negative examples relative to the positives (see 'Class imbalance' below for details).

We converted entity pairs into features by extracting: 1) the drug entity and the disorder entity 768-dimensional embeddings using the corresponding NER model—when entities consist of multiple tokens, we use the average of their token embeddings as the entity embedding—, 2) contextual information, i.e., [CLS] token 768-dimensional embeddings for the note, and 3) the 8-dimensional probability distribution predicted for the entities by the NER model. The 8 options the model must choose from are: '[PAD]', '[CLS]', 'X', 'O', 'I-Disorder', 'B-Disorder', 'I-Drug', and 'B-Drug'. Next, we explain each of these tokens. '[PAD]' and '[CLS]' are standard tokens in the LLM vocabulary: '[PAD]' is the padding token and '[CLS]' is the 'class token' commonly used in sequence classification problems. 'X' is a token used to train NER models and is used as the label of any *subword tokens* that are part of a named entity (e.g., the named entity 'paracetamol' may be split into tokens 'parac' and '#etamol' by the model tokenizer, and its labels would be 'B-Drug' and 'X', respectively; these are later combined during post-processing and only the label 'B-Drug' is relevant for the model predictions). Label 'O' is used to annotate tokens outside a drug or disorder mention. Labels 'I-Disorder' and 'B-Disorder' are the standard labels for inside a disorder concept and at the beginning of a disorder concept, respectively. Labels 'I-Drug' and 'B-Drug' have a similar meaning but for drug concepts.

These features (drug entity and disorder entity embeddings, contextual information, probability distribution) were concatenated to form a 3088-dimensional feature vector for each input entity pair. The RC features for the Bi-LSTM baseline model was different owing to differences in architecture and implementation. It used 300-D static word embeddings as input representation. The Bi-LSTM model does not have the '[PAD]', '[CLS]', 'X' labels as it works on word-level tokens without introducing special classification tokens or subword splitting tokens. In addition to the static word embeddings, we needed the contextualized representations, which we got from the Bi-LSTMs hidden state embedding, which were 256-D (128 each in both directions). Finally, the overall feature vector we create for the Bi-LSTM model contained 4*300+4*256+2*5= 2234 Dimensions, composed of two sets of static word and contextualized LSTM embeddings for each entity and similarly two sets of averaged embeddings for sentences to which each entity belongs for context, in addition to the 5-D NER probability vector for each entity

We trained the RC model in two settings: two-step and end-to-end. In the two-step setting, we used the gold-standard entity annotations from the Dutch ADE corpus. In the end-to-end setting, we use the entities predicted by the NER model as input to the RC classifier. To include a relation label between two predicted entities in the end-to-end setting, we required that there was a non-empty overlap between the predicted and gold-standard drug and relation entities.

*Class imbalance*

To address class imbalance between ADE and no-ADE as well as prescribing indication and no- prescribing indication labels, we experimented with various resampling strategies and sampling ratios in RC models (always using only the first fold in our 5-fold cross-validation). Since we generate candidate pairs to cover 99.42% of all ADE relations in the gold-standard Dutch ADE corpus, we end up with a high number of negative pairs (0.68% of the generated pairs are actual ADEs). To address this class imbalance, we experimented with the following techniques:

- Weighted random sampling (random sampling with per-class weighted probabilities)
- Synthetic minority oversampling technique (SMOTE) with random undersampling of majority class, with different final class ratios {0.25, 0.4, 0.5, 0.6, 0.75, 0.9}
- Adaptive synthetic sampling with random undersampling of majority class, with different final class ratios {0.25, 0.4, 0.5, 0.6, 0.75, 0.9}

The best performing resampling method was SMOTE with random undersampling with a final class ratio of 0.4, which we used in all experiments.

*Model training*

We trained separate binary classifiers for RC: one for ADE vs. no-ADE and one for prescribing indication vs. no- prescribing indication. We also used 5-fold cross validation for training RC models, where we use the corresponding NER models for each fold to generate entity pairs and their features. The RC model is a simple four-layer multi-layer perceptron (MLP) model. This is a simple and efficient approach since we did not train any Transformer-based model for RC and reuse the previously trained NER model for feature extraction of entity pairs. For RC, we conducted hyperparameter tuning for each fold and the search space was:

- Learning rate: {1e-5, 5e-5, 1e-6, 2e-6, 5e-6}
- Batch size: {32, 64, 128, 256, 512}
- Dropout: {0.2, 0.4, 0.5, 0.6, 0.8}
- Early stopping patience: {10, 20, 30}

The best hyperparameter setting for end-to-end RC was learning rate 1e-6, batch size 128, dropout 0.5, and patience 20. We also experimented with the number of model layers and the loss function (binary cross-entropy or focal loss). We found a 4-layer MLP with focal loss to be the best performing option.

*Model selection*

We benchmarked model selection strategies using F1 score on validation data, F2 score on validation data, and loss on validation data. F2-based models often had higher recall, as we show in our experiments.

*Threshold calculation*

To find the optimum threshold for models trained on each fold, we plotted the precision-recall curve (AUC-PR) on the validation set across all 5 folds (see Figure S1.1 below where we plot this curve for the MedRoBERTa.nl model). We then calculate the F1 and F2 scores for various thresholds and pick the thresholds that gave the highest score (F1 or F2). We then used these thresholds while inferring and benchmarking on the test set (internal evaluation) and on the external data.

*ADE group experiments*

As part of these experiments, we have mappings of drug-disorder pairs which belong to the same ADE from a clinical standpoint (i.e., ADE group). We evaluate model performance in two ways, where in order to consider a prediction a success the model must identify: 1) at least one ADE annotation part of an ADE group (the 'easy' setting) and 2) all ADE annotations part of an ADE group (the 'hard' setting).

We chose two thresholds for RC models: one that maximizes the F1 score on the validation set and another that maximizes the F2 score on the validation set. We predict with the RC model using both thresholds in our experiments reported in Table 3 and Table 4 in the main manuscript.

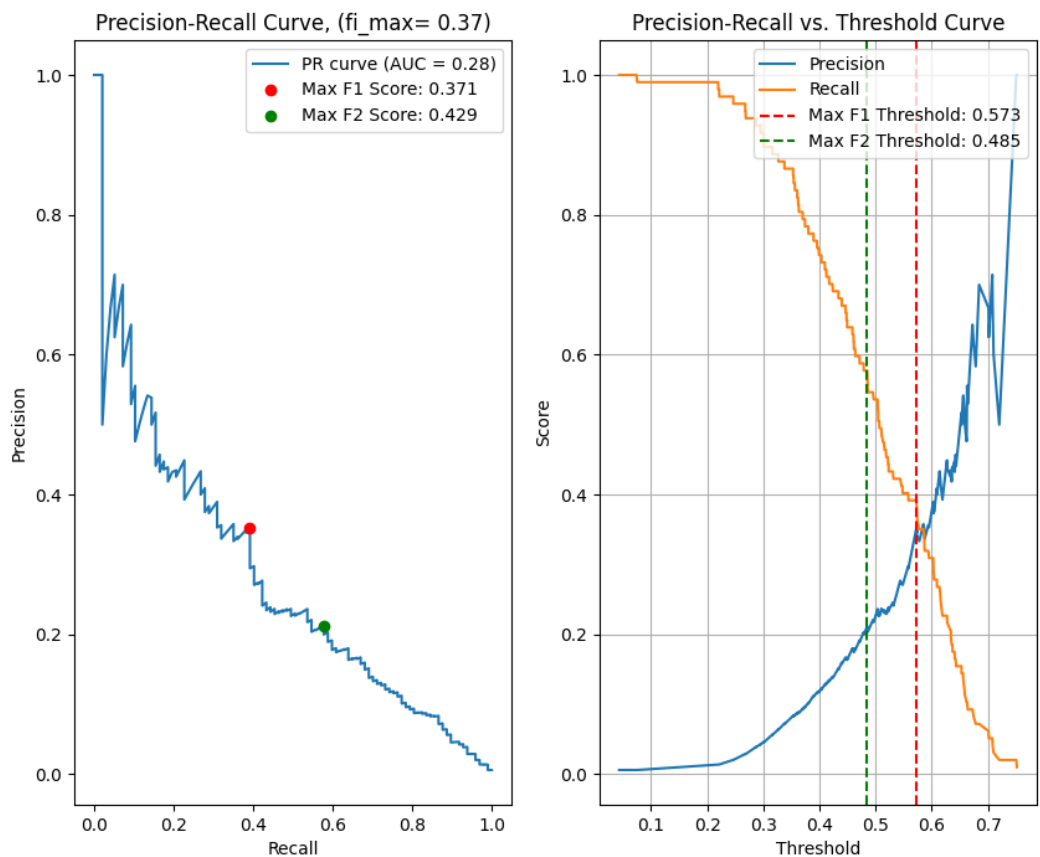

**Fig S1.2** On the left, the precision-recall curve for MedRoBERTa.nl on the Dutch ADE corpus' validation set (and the highest F1 and F2 scores we found). On the right, the precision and recall versus the threshold that maximizes F1 and F2 scores.

*ICU AKI and WINGS dataset experiments*

We conduct experiments using the ICU AKI and WINGS datasets as further internal and external validation of the models we investigate, respectively. We benchmarked two RC models in the end-to-end setting on the ICU AKI and WINGS datasets. We measured the precision, recall, specificity, and F1 score by taking both sets of thresholds (optimized F1 and F2). These metrics were calculated at the clinical note level, where a clinical note was labeled positive for ADE if even one drug-disorder pair in the document was labeled as ADE by our model. We first used our NER model to label drug and disorder entities in the note. We then created viable pairs of drug-disorder entities for the RC model the same way as we did for other experiments, as described in the "*Relation Classification (RC) models*" section above. We then used the ADE classification model to predict the probability of an ADE for all these entity pairs. We finally used the F1 and F2 thresholds for the corresponding models (computed using the internal validation set) to assign the positive/negative label and calculate the number of documents with at least one predicted ADE. Results for these experiments are reported in Table 3 and Table 4 in the main manuscript.

# Detection of Adverse Drug Events in Dutch clinical free text documents using Transformer Models – benchmark study

Supplement 2: Results

## Authors

Rachel M. Murphy*, PhD [a,b,c]

Nishant Mishra*, MSc [a,d]

Nicolette F. de Keizer, PhD [a,b,c]

Dave A. Dongelmans, MD PhD [c,e]

Kitty J. Jager, MD PhD [a,c,f]

Ameen Abu-Hanna, PhD [a,d,g]

Joanna E. Klopotowska**, PharmD PhD [a,b,c]  (corresponding author, j.e.klopotowska@amsterdamumc.nl)

Iacer Calixto**, PhD [a,d,g]

*shared first authors*

**shared last authors*

[a] Amsterdam UMC location University of Amsterdam, Department of Medical Informatics, Meibergdreef 9, Amsterdam, the Netherlands

[b] Amsterdam Public Health, Digital Health, Amsterdam, the Netherlands

[c] Amsterdam Public Health, Quality of Care, Amsterdam, the Netherlands

[d] Amsterdam Public Health, Methodology, Amsterdam, the Netherlands

[e] Amsterdam UMC location University of Amsterdam, Department of Intensive Care Medicine, Meibergdreef 9, Amsterdam, the Netherlands

[f] Amsterdam Public Health, Aging & Later Life, Amsterdam, the Netherlands

[g] Amsterdam Public Health, Mental Health, Amsterdam, the Netherlands

**Table S2.1** Named entity recognition models on drug and disorder entities on the test set of the Dutch ADE corpus (5-fold cross-validation).*

|  | Drug | | | Disorder | | | Overall | | |
|---|---|---|---|---|---|---|---|---|---|
| Model | Precision Micro | Recall Micro | F1 score Micro | Precision Micro | Recall Micro | F1 score Micro | Precision Micro | Recall Micro | F1 score Micro |
| | **Strict matching** | | | | | | | | |
| BiLSTM | 0.9235 ± 0.0075 | 0.7862 ± 0.0261 | 0.8491 ± 0.0140 | 0.7386 ± 0.0383 | 0.6075 ± 0.0249 | 0.6664 ± 0.0265 | 0.8048 ± 0.0268 | 0.6710 ± 0.0212 | 0.7318 ± 0.0222 |
| BERTje | 0.8591 ± 0.0136 | 0.8975 ± 0.0035 | 0.8778 ± 0.0085 | 0.6990 ± 0.0192 | 0.7238 ± 0.0276 | 0.7110 ± 0.0208 | 0.7575 ± 0.0153 | 0.7870 ± 0.0172 | 0.7719 ± 0.0140 |
| RobBERT-base | 0.8910 ± 0.0180 | 0.9097 ± 0.0064 | 0.9001 ± 0.0093 | 0.7347 ± 0.0139 | 0.7622 ± 0.0279 | 0.7480 ± 0.0197 | 0.7907 ± 0.0110 | 0.8160 ± 0.0159 | 0.8031 ± 0.0120 |
| MedRoBERTA.nl | **0.9284 ± 0.0216** | **0.9344 ± 0.0194** | **0.9311 ± 0.0075** | **0.7685 ± 0.0187** | **0.8088 ± 0.0244** | **0.7880 ± 0.0178** | **0.8244 ± 0.0109** | **0.8544 ± 0.0198** | **0.8390 ± 0.0130** |
| NuNER-multilingual | 0.9001 ± 0.0140 | 0.9113 ± 0.0144 | 0.9055 ± 0.0082 | 0.7515 ± 0.0163 | 0.7580 ± 0.0266 | 0.7547 ± 0.0211 | 0.8056 ± 0.0090 | 0.8138 ± 0.0198 | 0.8096 ± 0.0133 |
| | **Lenient matching** | | | | | | | | |
| BiLSTM | 0.9344 ± 0.0101 | 0.7954 ± 0.0245 | 0.8591 ± 0.0125 | 0.8801 ± 0.0240 | 0.7242 ± 0.0203 | 0.7942 ± 0.0118 | **0.8993 ± 0.0148** | 0.7499 ± 0.0117 | 0.8177 ± 0.0092 |
| BERTje | 0.8732 ± 0.0115 | 0.9122 ± 0.0032 | 0.8922 ± 0.0065 | 0.8193 ± 0.0093 | 0.8483 ± 0.0216 | 0.8334 ± 0.0100 | 0.8389 ± 0.0082 | 0.8716 ± 0.0129 | 0.8549 ± 0.0055 |
| RobBERT-base | 0.9038 ± 0.0180 | 0.9227 ± 0.0057 | 0.9131 ± 0.0089 | 0.8447 ± 0.0048 | 0.8761 ± 0.0188 | 0.8603 ± 0.0082 | 0.8656 ± 0.0060 | 0.8932 ± 0.0095 | 0.8791 ± 0.0039 |
| MedRoBERTA.nl | **0.9392 ± 0.0214** | **0.9452 ± 0.0187** | **0.9419 ± 0.0055** | **0.8658 ± 0.0183** | **0.9111 ± 0.0157** | **0.8877 ± 0.0104** | 0.8913 ± 0.0109 | **0.9236 ± 0.0134** | **0.9070 ± 0.0073** |
| NuNER-multilingual | 0.9125 ± 0.0137 | 0.9238 ± 0.0113 | 0.9180 ± 0.0043 | 0.8606 ± 0.0046 | 0.8679 ± 0.0156 | 0.8642 ± 0.0090 | 0.8793 ± 0.0048 | 0.8882 ± 0.0131 | 0.8837 ± 0.0052 |

* all measures with one standard deviation.

**Table S2.2** 'Drug', 'disorder' and 'overall' entity recognition task on validation set.*

|  | Drug | | | Disorder | | | Overall | | |
|---|---|---|---|---|---|---|---|---|---|
| Model | Precision Micro | Recall Micro | F1 score Micro | Precision Micro | Recall Micro | F1 score Micro | Precision Micro | Recall Micro | F1 score Micro |
| | **Strict matching** | | | | | | | | |
| BiLSTM | **0.9286 ± 0.0128** | 0.7948 ± 0.0247 | 0.8564 ± 0.0188 | 0.7526 ± 0.0314 | 0.5988 ± 0.0125 | 0.6666 ± 0.0110 | 0.8196 ± 0.0195 | 0.6702 ± 0.0062 | 0.7373 ± 0.0079 |
| BERTje | 0.8664 ± 0.0384 | 0.9020 ± 0.0075 | 0.8834 ± 0.0176 | 0.7072 ± 0.0131 | 0.7182 ± 0.0128 | 0.7126 ± 0.0088 | 0.7672 ± 0.0215 | 0.7865 ± 0.0072 | 0.7765 ± 0.0105 |
| RobBERT-base | 0.9006 ± 0.0262 | 0.9037 ± 0.0147 | 0.9021 ± 0.0205 | 0.7387 ± 0.0180 | 0.7462 ± 0.0338 | 0.7421 ± 0.0208 | 0.7986 ± 0.0208 | 0.8046 ± 0.0235 | 0.8015 ± 0.0184 |
| MedRoBERTA.nl | 0.9264 ± 0.0125 | **0.9468 ± 0.0116** | **0.9364 ± 0.0092** | **0.7812 ± 0.0122** | **0.7953 ± 0.0270** | **0.7879 ± 0.0125** | **0.8353 ± 0.0090** | **0.8516 ± 0.0157** | **0.8432 ± 0.0046** |
| NuNER-multilingual | 0.9010 ± 0.0312 | 0.9245 ± 0.0134 | 0.9125 ± 0.0204 | 0.7685 ± 0.0113 | 0.7437 ± 0.0337 | 0.7557 ± 0.0222 | 0.8195 ± 0.0152 | 0.8103 ± 0.0249 | 0.8149 ± 0.0199 |
| | **Lenient matching** | | | | | | | | |
| BiLSTM | **0.9399 ± 0.0122** | 0.8045 ± 0.0248 | 0.8668 ± 0.0186 | **0.8889 ± 0.0203** | 0.7078 ± 0.0227 | 0.7876 ± 0.0064 | **0.9083 ± 0.0136** | 0.7428 ± 0.0126 | 0.8171 ± 0.0071 |
| BERTje | 0.8815 ± 0.0400 | 0.9177 ± 0.0068 | 0.8988 ± 0.0190 | 0.8328 ± 0.0160 | 0.8458 ± 0.0131 | 0.8391 ± 0.0095 | 0.8511 ± 0.0248 | 0.8725 ± 0.0077 | 0.8615 ± 0.0126 |
| RobBERT-base | 0.9133 ± 0.0257 | 0.9164 ± 0.0142 | 0.9148 ± 0.0199 | 0.8525 ± 0.0197 | 0.8609 ± 0.0307 | 0.8563 ± 0.0166 | 0.8749 ± 0.0212 | 0.8814 ± 0.0210 | 0.8780 ± 0.0162 |
| MedRoBERTA.nl | 0.9375 ± 0.0126 | **0.9582 ± 0.0111** | **0.9477 ± 0.0089** | 0.8861 ± 0.0171 | **0.9019 ± 0.0244** | **0.8936 ± 0.0088** | 0.9051 ± 0.0133 | **0.9227 ± 0.0147** | **0.9137 ± 0.0055** |
| NuNER-multilingual | 0.9119 ± 0.0273 | 0.9358 ± 0.0147 | 0.9236 ± 0.0174 | 0.8770 ± 0.0068 | 0.8485 ± 0.0272 | 0.8623 ± 0.0134 | 0.8906 ± 0.0096 | 0.8804 ± 0.0173 | 0.8854 ± 0.0126 |

* all measures with one standard deviation

**Table S2.3** ADE and indication binary relation classification models on validation set using F1 threshold*.

| Model | ADE vs. no-ADE | | | | | | Indication vs. no-indication | | | | | |
|---|---|---|---|---|---|---|---|---|---|---|---|---|
| | Micro precision | Micro recall | Micro F1 score | Macro precision | Macro recall | Macro F1 score | Micro precision | Micro recall | Micro F1 score | Macro precision | Macro recall | Macro F1 score |
| **Using gold standard entities (two-step task)** | | | | | | | | | | | | |
| BERTje | 0.9885 ± 0.0045 | 0.9885 ± 0.0045 | 0.9885 ± 0.0045 | 0.6288 ± 0.0370 | 0.6127 ± 0.0300 | 0.6175 ± 0.0292 | 0.9689 ± 0.0086 | 0.9689 ± 0.0086 | 0.9689 ± 0.0086 | 0.6068 ± 0.0378 | 0.6030 ± 0.0206 | 0.6049 ± 0.0081 |
| RobBERT | 0.9890 ± 0.0018 | 0.9890 ± 0.0018 | 0.9890 ± 0.0018 | 0.6246 ± 0.0230 | 0.6525 ± 0.0289 | 0.6348 ± 0.0217 | 0.9754 ± 0.0033 | 0.9754 ± 0.0033 | 0.9754 ± 0.0033 | 0.6534 ± 0.0308 | 0.6238 ± 0.0206 | 0.6325 ± 0.0073 |
| MedRoBERTA.nl | 0.9878 ± 0.0031 | 0.9878 ± 0.0031 | 0.9878 ± 0.0031 | 0.6247 ± 0.0342 | 0.6847 ± 0.0679 | 0.6448 ± 0.0165 | 0.9731 ± 0.0030 | 0.9731 ± 0.0030 | 0.9731 ± 0.0030 | 0.6403 ± 0.0240 | 0.6474 ± 0.0135 | 0.6422 ± 0.0145 |
| NuNER-multilingual | 0.9894 ± 0.0043 | 0.9894 ± 0.0043 | 0.9894 ± 0.0043 | 0.6371 ± 0.0652 | 0.6560 ± 0.0482 | 0.6406 ± 0.0266 | 0.9688 ± 0.0063 | 0.9688 ± 0.0063 | 0.9688 ± 0.0063 | 0.6092 ± 0.0243 | 0.6274 ± 0.0277 | 0.6141 ± 0.0189 |
| **Using predicted entities with lenient matching (end-to-end task)** | | | | | | | | | | | | |
| BERTje | 0.9900 ± 0.0025 | 0.9900 ± 0.0025 | 0.9900 ± 0.0025 | 0.6157 ± 0.0495 | 0.6490 ± 0.0307 | 0.6184 ± 0.0305 | 0.9728 ± 0.0084 | 0.9728 ± 0.0084 | 0.9728 ± 0.0084 | 0.6110 ± 0.0661 | 0.6224 ± 0.0108 | 0.6055 ± 0.0212 |
| RobBERT | 0.9905 ± 0.0018 | 0.9905 ± 0.0018 | 0.9905 ± 0.0018 | 0.6455 ± 0.0443 | 0.6625 ± 0.0495 | 0.6443 ± 0.0244 | 0.9775 ± 0.0047 | 0.9775 ± 0.0047 | 0.9775 ± 0.0047 | 0.6406 ± 0.0326 | 0.6313 ± 0.0197 | 0.6313 ± 0.0122 |
| MedRoBERTA.nl | 0.9910 ± 0.0030 | 0.9910 ± 0.0030 | 0.9910 ± 0.0030 | 0.6567 ± 0.0482 | 0.6565 ± 0.0677 | 0.6455 ± 0.0301 | 0.9778 ± 0.0012 | 0.9778 ± 0.0012 | 0.9778 ± 0.0012 | 0.6510 ± 0.0203 | 0.6393 ± 0.0210 | 0.6443 ± 0.0179 |
| NuNER-multilingual | 0.9904 ± 0.0031 | 0.9904 ± 0.0031 | 0.9904 ± | 0.6466 ± | 0.6514 ± | 0.6409 ± 0.0140 | 0.9783 ± 0.0053 | 0.9783 ± 0.0053 | 0.9783 ± 0.0053 | 0.6318 ± 0.0254 | 0.6195 ± 0.0195 | 0.6216 ± 0.0081 |

|  |  |  | 0.0031 | 0.0415 | 0.0395 |  |  |  |  |  |  |  |
|---|---|---|---|---|---|---|---|---|---|---|---|---|

* all measures with one standard deviation. Note: micro-averaged recall, precision and F-score are the same for binary classification.

**Table S2.4** ADE and indication binary relation classification models on validation set using F2 threshold.*

| Model | ADE vs. no-ADE | | | | | | Indication vs. no-indication | | | | | |
|---|---|---|---|---|---|---|---|---|---|---|---|---|
| | Micro precision | Micro recall | Micro F2 score | Macro precision | Macro recall | Macro F2 score | Micro precision | Micro recall | Micro F2 score | Macro precision | Macro recall | Macro F2 score |
| **Using gold standard entities (two-step task)** | | | | | | | | | | | | |
| BERTje | 0.9733 ± 0.0143 | 0.9700 ± 0.0129 | 0.9700 ± 0.0129 | 0.5685 ± 0.0319 | 0.6986 ± 0.0347 | 0.6260 ± 0.0251 | 0.9262 ± 0.0107 | 0.9262 ± 0.0107 | 0.9262 ± 0.0107 | 0.5454 ± 0.0047 | 0.6605 ± 0.0170 | 0.5967 ± 0.0091 |
| RobBERT | 0.9818 ± 0.0038 | 0.9818 ± 0.0038 | 0.9818 ± 0.0038 | 0.5836 ± 0.0140 | 0.7057 ± 0.0342 | 0.6557 ± 0.0165 | 0.9467 ± 0.0150 | 0.9467 ± 0.0150 | 0.9467 ± 0.0150 | 0.5747 ± 0.0159 | 0.6841 ± 0.0201 | 0.6345 ± 0.0138 |
| MedRoBERTA.nl | 0.9791 ± 0.0036 | 0.9791 ± 0.0036 | 0.9791 ± 0.0036 | 0.5822 ± 0.0085 | 0.7541 ± 0.0444 | 0.6757 ± 0.0225 | 0.9527 ± 0.0087 | 0.9527 ± 0.0087 | 0.9527 ± 0.0087 | 0.5882 ± 0.0176 | 0.7022 ± 0.0098 | 0.6551 ± 0.0109 |
| NuNER-multilingual | 0.9849 ± 0.0086 | 0.9849 ± 0.0086 | 0.9849 ± 0.0086 | 0.6068 ± 0.0378 | 0.7295 ± 0.0335 | 0.6525 ± 0.0189 | 0.9317 ± 0.0187 | 0.9317 ± 0.0187 | 0.9317 ± 0.0187 | 0.5639 ± 0.0236 | 0.6977 ± 0.0239 | 0.6250 ± 0.0207 |
| **Using predicted entities with lenient matching (end-to-end task)** | | | | | | | | | | | | |
| BERTje | 0.9797 ± 0.0089 | 0.9797 ± 0.0089 | 0.9797 ± 0.0089 | 0.5723 ± 0.0277 | 0.7160 ± 0.0439 | 0.6411 ± 0.0159 | 0.9500 ± 0.0187 | 0.9500 ± 0.0187 | 0.9500 ± 0.0187 | 0.5663 ± 0.0267 | 0.6798 ± 0.0356 | 0.6190 ± 0.0092 |
| RobBERT | 0.9803 ± 0.0053 | 0.9803 ± 0.0053 | 0.9803 ± 0.0053 | 0.5820 ± 0.0190 | 0.7528 ± 0.0464 | 0.6736 ± 0.0280 | 0.9566 ± 0.0019 | 0.9566 ± 0.0019 | 0.9566 ± 0.0019 | 0.5747 ± 0.0038 | 0.6959 ± 0.0139 | 0.6443 ± 0.0078 |
| MedRoBERTA.nl | 0.9787 ± 0.0049 | 0.9787 ± 0.0049 | 0.9787 ± 0.0049 | 0.5835 ± 0.0195 | 0.7958 ± 0.0405 | 0.6906 ± 0.0330 | 0.9553 ± 0.0052 | 0.9553 ± 0.0052 | 0.9553 ± 0.0052 | 0.5842 ± 0.0080 | 0.7142 ± 0.0152 | 0.6597 ± 0.0112 |
| NuNER-multilingual | 0.9838 ± 0.0100 | 0.9838 ± 0.0100 | 0.9838 ± 0.0100 | 0.5706 ± 0.0182 | 0.7670 ± 0.0237 | 0.6627 ± 0.0243 | 0.9531 ± 0.0070 | 0.9531 ± 0.0070 | 0.9531 ± 0.0070 | 0.5665 ± 0.0063 | 0.6955 ± 0.0202 | 0.6357 ± 0.0112 |

*all measures with one standard deviation. Note: Micro-averaged recall, precision and F-score are the same for binary classification.

**Table S2.5** ADE and indication binary relation classification models on test set using F1 threshold.*

| Model | ADE vs. no-ADE | | | | | | Indication vs. no-indication | | | | | |
|---|---|---|---|---|---|---|---|---|---|---|---|---|
| | Micro precision | Micro recall | Micro F1 score | Macro precision | Macro recall | Macro F1 score | Micro precision | Micro recall | Micro F1 score | Macro precision | Macro recall | Macro F1 score |
| **Using gold standard entities (two-step task)** | | | | | | | | | | | | |
| BERTje | 0.9880 ± 0.0039 | 0.9880 ± 0.0039 | 0.9880 ± 0.0039 | 0.6139 ± 0.0462 | 0.6010 ± 0.0205 | 0.6005 ± 0.0146 | 0.9680 ± 0.0045 | 0.9680 ± 0.0045 | 0.9680 ± 0.0045 | 0.5927 ± 0.0129 | 0.5962 ± 0.0220 | 0.5925 ± 0.0135 |
| RobBERT | 0.9879 ± 0.0030 | 0.9879 ± 0.0030 | 0.9879 ± 0.0030 | 0.6133 ± 0.0251 | 0.6290 ± 0.0384 | 0.6181 ± 0.0271 | 0.9744 ± 0.0037 | 0.9744 ± 0.0037 | 0.9744 ± 0.0037 | 0.6255 ± 0.0176 | 0.6086 ± 0.0287 | 0.6105 ± 0.0168 |
| MedRoBERTA.nl | 0.9847 ± 0.0049 | 0.9847 ± 0.0049 | 0.9847 ± 0.0049 | 0.6030 ± 0.0283 | 0.6631 ± 0.0433 | 0.6301 ± 0.0196 | 0.9691 ± 0.0070 | 0.9691 ± 0.0070 | 0.9691 ± 0.0070 | 0.6258 ± 0.0333 | 0.6285 ± 0.0286 | 0.6202 ± 0.0110 |
| NuNER-multilingual | 0.9892 ± 0.0034 | 0.9892 ± 0.0034 | 0.9892 ± 0.0034 | 0.6218 ± 0.0312 | 0.6001 ± 0.0429 | 0.6133 ± 0.0315 | 0.9700 ± 0.0045 | 0.9700 ± 0.0045 | 0.9700 ± 0.0045 | 0.6040 ± 0.0229 | 0.6190 ± 0.0216 | 0.6082 ± 0.0131 |
| **Using predicted entities with lenient matching (end-to-end task)** | | | | | | | | | | | | |
| BERTje | 0.9846 ± 0.0072 | 0.9846 ± 0.0072 | 0.9846 ± 0.0072 | 0.5982 ± 0.0434 | 0.6219 ± 0.0418 | 0.5963 ± 0.0160 | 0.9694 ± 0.0036 | 0.9694 ± 0.0036 | 0.9694 ± 0.0036 | 0.6014 ± 0.0148 | 0.5967 ± 0.0201 | 0.5960 ± 0.0100 |
| RobBERT | 0.9884 ± 0.0024 | 0.9884 ± 0.0024 | 0.9884 ± 0.0024 | 0.6123 ± 0.0280 | 0.6096 ± 0.0446 | 0.6040 ± 0.0193 | 0.9738 ± 0.0043 | 0.9738 ± 0.0043 | 0.9738 ± 0.0043 | 0.6413 ± 0.0358 | 0.5967 ± 0.0220 | 0.6088 ± 0.0106 |
| MedRoBERTA.nl | 0.9891 ± 0.0025 | 0.9891 ± 0.0025 | 0.9891 ± 0.0025 | 0.6300 ± 0.0247 | 0.6209 ± 0.0210 | 0.6222 ± 0.0094 | 0.9728 ± 0.0063 | 0.9728 ± 0.0063 | 0.9728 ± 0.0063 | 0.6440 ± 0.0380 | 0.6097 ± 0.0247 | 0.6180 ± 0.0120 |
| NuNER-multilingual | 0.9896 ± 0.0026 | 0.9896 ± 0.0026 | 0.9896 ± 0.0026 | 0.6349 ± 0.0502 | 0.5957 ± 0.0289 | 0.6041 ± 0.0251 | 0.9753 ± 0.0013 | 0.9753 ± 0.0013 | 0.9753 ± 0.0013 | 0.6461 ± 0.0201 | 0.5996 ± 0.0109 | 0.6173 ± 0.0116 |

* all measures with one standard deviation. Note: Micro-averaged recall, precision and F-score are the same for binary classification.

**Table S2.6** ADE and indication binary relation classification models on test set using F2 threshold.*

| Model | ADE vs. no-ADE | | | | | | Indication vs. no-indication | | | | | |
|---|---|---|---|---|---|---|---|---|---|---|---|---|
| | Micro precision | Micro recall | Micro F2 score | Macro precision | Macro recall | Macro F2 score | Micro precision | Micro recall | Micro F2 score | Macro precision | Macro recall | Macro F2 score |
| **Using gold standard entities (two-step task)** | | | | | | | | | | | | |
| BERTje | 0.9695 ± 0.0101 | 0.9695 ± 0.0101 | 0.9695 ± 0.0101 | 0.5504 ± 0.0107 | 0.6815 ± 0.0214 | 0.6131 ± 0.0147 | 0.9280 ± 0.0130 | 0.9280 ± 0.0130 | 0.9280 ± 0.0130 | 0.5486 ± 0.0058 | 0.6587 ± 0.0360 | 0.5995 ± 0.0150 |
| RobBERT | 0.9793 ± 0.0074 | 0.9793 ± 0.0074 | 0.9793 ± 0.0074 | 0.5715 ± 0.0204 | 0.6649 ± 0.0316 | 0.6277 ± 0.0290 | 0.9473 ± 0.0131 | 0.9473 ± 0.0131 | 0.9473 ± 0.0131 | 0.5774 ± 0.0186 | 0.6729 ± 0.0264 | 0.6313 ± 0.0140 |
| MedRoBERTA.nl | 0.9752 ± 0.0063 | 0.9752 ± 0.0063 | 0.9752 ± 0.0063 | 0.5685 ± 0.0137 | 0.7159 ± 0.0457 | 0.6449 ± 0.0216 | 0.9455 ± 0.0137 | 0.9455 ± 0.0137 | 0.9455 ± 0.0137 | 0.5772 ± 0.0182 | 0.6826 ± 0.0140 | 0.6360 ± 0.0144 |
| NuNER-multilingual | 0.9760 ± 0.0096 | 0.9760 ± 0.0096 | 0.9760 ± 0.0096 | 0.5638 ± 0.0195 | 0.6678 ± 0.0237 | 0.6207 ± 0.0210 | 0.9362 ± 0.0102 | 0.9362 ± 0.0102 | 0.9362 ± 0.0102 | 0.5610 ± 0.0142 | 0.6728 ± 0.0119 | 0.6186 ± 0.0180 |
| **Using predicted entities with lenient matching (end-to-end task)** | | | | | | | | | | | | |
| BERTje | 0.9772 ± 0.0076 | 0.9772 ± 0.0076 | 0.9772 ± 0.0076 | 0.5650 ± 0.0126 | 0.6696 ± 0.0172 | 0.6238 ± 0.0093 | 0.9495 ± 0.0120 | 0.9495 ± 0.0120 | 0.9495 ± 0.0120 | 0.5635 ± 0.0041 | 0.6336 ± 0.0361 | 0.6040 ± 0.0162 |
| RobBERT | 0.9780 ± 0.0074 | 0.9780 ± 0.0074 | 0.9780 ± 0.0074 | 0.5683 ± 0.0198 | 0.6704 ± 0.0272 | 0.6273 ± 0.0253 | 0.9538 ± 0.0080 | 0.9538 ± 0.0080 | 0.9538 ± 0.0080 | 0.5792 ± 0.0079 | 0.6566 ± 0.0181 | 0.6277 ± 0.0059 |
| MedRoBERTA.nl | 0.9713 ± 0.0150 | 0.9713 ± 0.0150 | 0.9713 ± 0.0150 | 0.5719 ± 0.0236 | 0.7318 ± 0.0477 | 0.6450 ± 0.0225 | 0.9465 ± 0.0097 | 0.9465 ± 0.0097 | 0.9465 ± 0.0097 | 0.5762 ± 0.0140 | 0.6828 ± 0.0164 | 0.6374 ± 0.0148 |
| NuNER-multilingual | 0.9762 ± 0.0102 | 0.9762 ± 0.0102 | 0.9762 ± 0.0102 | 0.5758 ± 0.0330 | 0.6912 ± 0.0179 | 0.6383 ± 0.0303 | 0.9508 ± 0.0112 | 0.9508 ± 0.0112 | 0.9508 ± 0.0112 | 0.5771 ± 0.0248 | 0.6513 ± 0.0154 | 0.6206 ± 0.0149 |

* all measures with one standard deviation. Note: Micro-averaged recall, precision and F-score are the same for binary classification.